\documentclass[journal,10pt,twoside]{IEEEtran}

\usepackage{amsmath,amssymb,amsthm}
\usepackage{bm}           
\usepackage{booktabs}     
\usepackage{subcaption}   
\usepackage{graphicx}
\usepackage[linesnumbered,ruled,vlined]{algorithm2e} 
\usepackage{tikz}
\usetikzlibrary{arrows.meta,positioning}
\usepackage{xcolor}
\usepackage{hyperref}
\usepackage{cite}
\usepackage{url}
\usepackage{multirow}
\usepackage{array}
\usepackage{multicol}
\usepackage{tabularx}
\usepackage{overpic}
\definecolor{myblue}{RGB}{31,119,180}
\definecolor{myteal}{RGB}{15,110,86}
\definecolor{myorange}{RGB}{186,117,23}
\definecolor{mycoral}{RGB}{216,90,48}
\definecolor{mypurple}{RGB}{127,119,221}

\newcommand{\X}{\mathcal{X}}     
\newcommand{\LL}{\mathcal{L}}    
\newcommand{\Z}{\mathcal{Z}}     
\newcommand{\D}{\mathcal{D}}     
\newcommand{\w}{w_{tkj}}         

\newcommand{\muu}{\boldsymbol{\mu}}
\newcommand{\Sig}{\boldsymbol{\Sigma}}
\newcommand{\xt}{\mathbf{x}_t}
\newcommand{\lj}{\ell_j}
\newcommand{\dtk}{d_{tk}}
\newcommand{\ztk}{z_{tk}}
\newcommand{\R}{\mathbb{R}}
\DeclareMathOperator*{\argmax}{arg\,max}

\newcommand{\bGamma}{\boldsymbol{\Gamma}}
\newtheorem{assumption}{Assumption}

\usepackage{orcidlink}
\hypersetup{
  colorlinks=true,
  linkcolor=myblue,
  citecolor=myteal,
  urlcolor=myblue,
  pdftitle={BPDA-GMM: Bayesian Probabilistic Data Association via GMM for Semantic SLAM},
}

\begin{document}

\title{BPDA-GMM: Bayesian Probabilistic Data Association via Gaussian Mixture Models for Semantic SLAM}
\author{Thanh Nguyen Canh\orcidlink{https://orcid.org/0000-0001-6332-1002},~\IEEEmembership{Graduate Student Member,~IEEE}, Haolan Zhang\orcidlink{https://orcid.org/0000-0001-6332-1002},~\IEEEmembership{Graduate Student Member,~IEEE}, Xiem HoangVan\orcidlink{https://orcid.org/0000-0001-6332-1002}, Antonio Sgorbissa\orcidlink{0000-0001-7789-4311} and Nak Young Chong\orcidlink{https://orcid.org/0000-0001-5736-0769},~\IEEEmembership{Senior Member,~IEEE} 
\thanks{This work was supported by JST SPRING, Japan Grant Number JPMJSP2102. (Corresponding authors: nakyoung@jaist.ac.jp, xiemhoang@vnu.edu.vn)}
\thanks{Thanh Nguyen Canh, Haolan Zhang, and Nak Yong Chong are School of Information Science, Japan Advanced Institute of Science and Technology,
Ishikawa, 923-1211, Japan.
        {\tt\small \{thanhnc, haolan.z, nakyoung\}@jaist.ac.jp}}%
\thanks{Thanh Nguyen Canh and Xiem HoangVan are with the University of Engineering and Technology, Vietnam National University, Hanoi, 10000, Vietnam. {\tt\small \{xiemhoang\}@vnu.edu.vn}}
}


\maketitle

\begin{abstract}
Probabilistic data association (PDA) improves semantic SLAM in perceptually aliased scenes, but existing methods often assume a fixed landmark set, recompute association weights as the map grows, or rely on hand-tuned null-hypothesis weights. To address these limitations, we propose \textbf{BPDA-GMM}, an online Bayesian PDA framework for semantic SLAM with a growing object-level map. BPDA-GMM uses a Dirichlet-process prior to induce a Chinese Restaurant Process (CRP) association model, where accumulated evidence favors existing landmarks, and the concentration parameter assigns probability mass to new landmarks. For each semantic detection, plausible candidates are selected by a joint semantic-geometric gate, CRP-weighted association probabilities are computed, and object landmarks are updated as semantic Gaussians in closed form. The resulting landmark set forms a Gaussian mixture model, and its dominant component is passed to the back-end as a max-mixture semantic factor. When association weights are inconclusive, an ambiguity-triggered $\alpha$-divergence tempering step improves discrimination. Finally, a decoupled back-end zeroes the pose Jacobian of semantic factors, allowing noisy detections to refine landmarks without directly perturbing the trajectory. Experiments in simulation and on a real indoor dataset demonstrate improved trajectory accuracy, semantic mapping quality, and robustness to perceptual aliasing and classifier errors over state-of-the-art baselines. Code and video are publicly available at \url{https://anonymous.4open.science/r/BPDA-SLAM-8449/}.
\end{abstract}


\section{Introduction}
\label{sec:intro}

\IEEEPARstart{S}{emantic} simultaneous localization and mapping (SLAM) augments metric SLAM by associating geometric landmarks with semantic labels. By representing the environment in terms of object-level entities, semantic SLAM enables robots to build human-interpretable maps, reason about scene structure, and use semantic landmarks for navigation, manipulation, and loop closure~\cite{canh2025semantic}. However, these benefits depend strongly on reliable data association. When multiple landmarks share the same semantic class, the robot must decide which object generated each observation. Maximum-likelihood association is efficient, but it makes a hard assignment for each observation.  If this assignment is incorrect, it is inserted into the back-end as a biased constraint that can be difficult to correct later~\cite{brox2021maximum,bowman2017probabilistic}.

Existing PDA formulations for SLAM either iterate expectation-maximization~(EM) over the full keyframe history~\cite{bowman2017probabilistic}, freeze association weights at observation time via max-marginalization~\cite{olson2013inference, doherty2020probabilistic}, approximate the matrix permanent through ranked assignment~\cite{michael2022probabilistic}, or maintain explicit hypothesis trees over discrete association sequences~\cite{wang2018robust, hsiao2019mh}. However, these formulations share a common parametric assumption that assumes a fixed or effectively fixed landmark set. This assumption is poorly suited to online semantic SLAM, where the map grows as new objects are observed. As a result, prior PDA formulations often require recomputation or approximation of past association weights, manually tuned null-hypothesis probabilities, or expensive hypothesis management.


To address these limitations, we propose BPDA-GMM, an online Bayesian PDA framework for semantic SLAM. BPDA-GMM models association using a Dirichlet-process prior to induce a Chinese Restaurant Process (CRP)~\cite{kim2012modeling} association model, where accumulated evidence favors existing landmarks and the concentration parameter assigns probability mass to new ones. For each semantic detection, plausible candidates are selected by a joint semantic-geometric gate, and their likelihoods are combined with the CRP prior to compute association probabilities. Object landmarks are then updated as semantic Gaussians, forming a Gaussian mixture model whose dominant component is converted into a max-mixture semantic factor. To prevent noisy detections from directly perturbing the trajectory, the back-end zeroes the pose Jacobian of semantic factors, while semantic loop closure is handled through CRP-based landmark re-identification and merging.
The main contributions of this paper are summarized as follows:
\begin{enumerate}
    \item An \textbf{online Bayesian probabilistic data-association model} based on a Dirichlet-process prior, whose CRP representation supports a growing landmark set, incrementally accumulates association evidence, and assigns principled probability mass to new landmarks.

    \item A \textbf{semantic Gaussian mixture mapping framework} in which CRP-weighted observations update object landmarks as semantic Gaussians in closed form, forming a Gaussian mixture model.

    \item A \textbf{robust semantic SLAM optimization pipeline} that combines semantic-geometric gating, ambiguity-triggered posterior tempering, CRP-based landmark re-identification, and a decoupled back-end with zero pose Jacobians to improve robustness against perceptual aliasing and classifier errors.
\end{enumerate}

\section{Related Work}
\label{sec:related}

\subsection{Probabilistic Data Association in SLAM}

Probabilistic data association originated in target tracking~\cite{bar1975tracking, reid2003algorithm} and was adapted to SLAM by Cox and Leonard~\cite{cox1994modeling}. In semantic SLAM, Bowman \textit{et al.}~\cite{bowman2017probabilistic} formulated PDA as EM jointly over poses, landmark position, and association weights, but this formulation requires iterating over all previous measurements at each new keyframe. Doherty \textit{et al.} reduced this cost through multimodal semantic SLAM with nonparametric belief propagation~\cite{doherty2019multimodal}, but the resulting Gibbs sampling remains computationally expensive. Their later max-marginal formulation~\cite{doherty2020probabilistic} converts association uncertainty into a max-mixture factor~\cite{olson2013inference}, making it compatible with standard nonlinear least-squares SLAM, but the association weights are frozen at observation time.
Michael \textit{et al.}~\cite{michael2022probabilistic} approximated the \#P-complete matrix permanent underlying EM via Murty's ranked-assignment algorithm~\cite{murty1968algorithm}, achieving fast performance on small candidate sets while still inheriting the fixed-$M$ assumption. Multi-hypothesis methods~\cite{wang2018robust, hsiao2019mh} maintain explicit discrete association tracks. MHJCBB~\cite{wang2018robust} extends JCBB~\cite{neira2002data} with $K$ pruned parallel hypotheses, while MH-iSAM2~\cite{hsiao2019mh} incrementalizes multi-hypothesis inference through a Hypo-tree extension of the Bayes tree.

However, prior PDA formulations either fix $M$ or maintain explicit discrete hypothesis sets requiring $O(K)$ or exponential storage. BPDA-GMM marginalizes over associations implicitly through a Dirichlet-process prior whose CRP counts update in $O(1)$ per observation, requiring a single back-end pass per keyframe and treating new-landmark birth as a calibrated probability rather than a binary decision.

\subsection{Semantic Landmark Representation}
In parallel to PDA-based association methods, semantic SLAM also studies how semantic landmarks should be represented geometrically. CubeSLAM~\cite{yang2019cubeslam} models objects as 3D cuboids inferred from 2D bounding boxes, while QuadricSLAM~\cite{nicholson2019quadricslam} represents objects using constrained dual quadrics with compact size, position, and orientation parameters. Qian \textit{et al.}~\cite{qian2021semantic} extended quadric-based SLAM with bag-of-words bipartite matching for object-level association, and DSP-SLAM~\cite{wang2021dspslam} replaced algebraic shape models with category-specific deep shape embeddings for dense reconstruction. EAO-SLAM~\cite{wu2020eao} combines parametric and nonparametric statistical tests for ensemble object association, and its later extension~\cite{wu2023object} adds topological mapping for higher-level tasks. These methods provide structured object representations, but they typically commit to a single shape prior and resolve association through hard matching or fixed ensemble rules.

Gaussian mixture representations offer a more flexible alternative. SGBA~\cite{ji2024sgba} models LiDAR environments as semantic Gaussian mixture models without predefined feature types and uses a semantic-indicator constraint to restrict associations to landmarks of matching class. BPDA-GMM adopts this semantic consistency principle but extends it to online semantic SLAM with Bayesian probabilistic association and a growing landmark set. SlideSLAM~\cite{liu2025slideslam} also uses object-level semantic landmarks and introduces a zero-pose-Jacobian factor design, but it relies on hard data association and lacks a probabilistic null hypothesis. BPDA-GMM combines these complementary ideas: semantic landmarks are maintained as Gaussian components in a mixture map, associations are inferred through CRP-weighted probabilities, and the dominant mixture component is passed to a decoupled max-mixture back-end.

\section{Problem Formulation}
\label{sec:problem}

Consider a robot moving through an unknown environment populated by static semantic landmarks. The robot state at time~$t$ is $\xt \in SE(3)$, with trajectory $\X = \{\xt\}_{t=1}^T$. The map consists of $M$ landmarks $\LL = \{\lj\}_{j=1}^M$, each with a positional component $\lj^p \in \R^3$ and a semantic class $\lj^s \in \mathcal{C} = \{1,\ldots,C\}$. The landmark count $M$ is itself unknown and grows online as new objects are observed. At each keyframe $t$, the robot receives odometry $u_t \in SE(3)$ with covariance $\Sigma_t$, together with a set of semantic measurements $\Z_t = \{z_{tk}\}_{k=1}^{K_t}$, where $z_{tk} = (z_{tk}^p, z_{tk}^s)$ comprises a geometric centroid in the camera frame and a semantic score vector from an object detector. The complete SLAM problem is to estimate:
\begin{equation}
  \hat{\X},\hat{\LL} = \argmax_{\X,\LL}
    \; p(\X,\LL \mid \Z, \mathcal{U}).
  \label{eq:map}
\end{equation}

The posterior marginalizes over the unknown discrete data associations $\D = \{d_{tk}\}$,  with $d_{tk} = j$ means that $\ztk$ was generated by $\lj$ and $d_{tk} = 0$ denotes the null hypothesis, corresponding to either a false positive or a previously unseen landmark:
\begin{equation}
\small
  p(\X,\LL \mid \Z, \mathcal{U}) \propto
  \sum_{\D} p(\Z \mid \X, \LL, \D)\,
            p(\D)\,
            p(\X \mid \mathcal{U})\,
            p(\LL).
  \label{eq:marginal}
\end{equation}
\begin{assumption} \label{assump:for}
  Each measurement $\ztk$ is generated by at most one landmark, or is a false positive (null hypothesis).
\end{assumption}

\begin{assumption}\label{assump:inv}
  Measurements within a keyframe are conditionally independent given the robot pose and landmark states.
\end{assumption}


Exact evaluation of~\eqref{eq:marginal} is intractable. Marginalizing over $\D$ yields a Gaussian mixture whose summation is exponential in the number of measurements ($\#$P-complete in general via the matrix permanent). BPDA-GMM addresses this via an incremental EM scheme (Sec.~\ref{sec:method}) in which the E-step approximates $p(\D \mid \cdot)$ with Bayesian soft weights derived from the Dirichlet-process prior, and the M-step updates $(\X, \LL)$ through a decoupled landmark Kalman filter and a trajectory-only iSAM2 back-end.


\begin{figure*}[!t]
    \centering
    \includegraphics[width=0.89\textwidth]{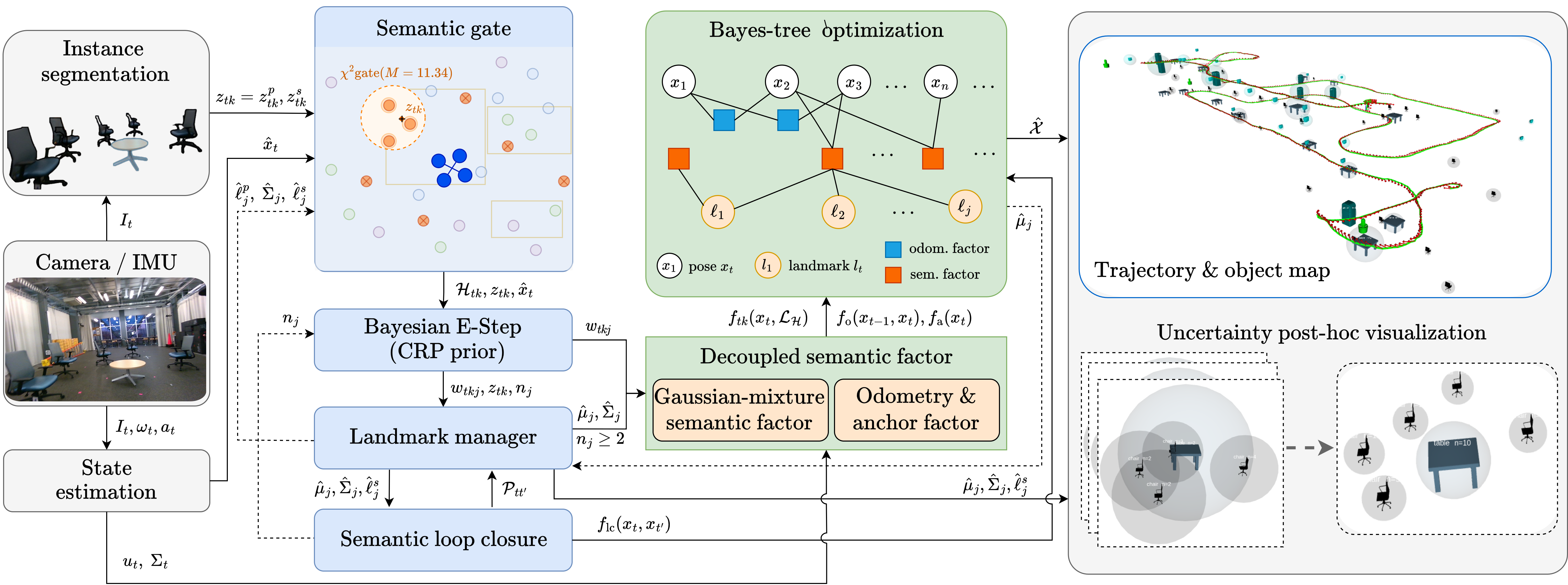}
    \caption{\textbf{BPDA-GMM system overview.} Each semantic detection is first filtered by a semantic-geometric gate, then assigned CRP-weighted $w_{tkj}$ association probabilities. These weights update semantic Gaussian landmarks $(\hat{\bm\mu}_j, \hat{\bm\Sigma}_j, \hat{\ell}^s_j)$ in the front-end, while the dominant mixture component is converted into a max-mixture semantic factor for the decoupled back-end.
    }
  \label{fig:overview}
\end{figure*}

\section{BPDA-GMM: Method}
\label{sec:method}
BPDA-GMM addresses the semantic SLAM problem by maintaining a Bayesian association posterior over a growing landmarks set and producing a single max-mixture factor per measurement. Fig.~\ref{fig:overview} shows the full pipeline, which includes six components, organized to mirror the posterior factorization in~\eqref{eq:marginal}. At the front-end, a Dirichlet process prior provides CRP counts as sufficient statistics that are updated in $O(1)$ (Sec.~\ref{sec:dp}); a semantic-and-geometric gate restricts each measurement to in-class, in-covariance candidates (Sec.~\ref{sec:likelihood}); and a Bayesian E-step computes association weights with $\alpha$-divergence tempering for noise robustness (Sec.~\ref{sec:estep}). The landmark layer updates per-landmark semantic Gaussians in closed form, with a merge rule that doubles as the loop-closure primitive (Sec.~\ref{sec:landmark}). At the back-end, a decoupled pose-landmark M-step isolates semantic measurement noise from trajectory estimation (Sec.~\ref{sec:mstep}). Semantic loop closure (Sec.~\ref{sec:loop_closure}) then follows from retaining dormant CRP counts and reusing the merge rule across place-matched keyframe pairs. Algorithm~\ref{alg:bpda} summarizes the per-keyframe pipeline.  

\subsection{Dirichlet Process Prior and Incremental Count}
\label{sec:dp}
To accommodate the unbounded landmark set, we place a \emph{Dirichlet process} prior on the association distribution:
\begin{equation}
  G \sim \mathrm{DP}(\alpha_0,\,G_0), \qquad
  \dtk \mid G \;\sim\; G,
  \label{eq:dp}\end{equation}
where $\alpha_0 > 0$ is the concentration parameter and $G_0$ defines the new-landmark base distribution.
Marginalizing out $G$ yields the CRP prior conditioned on all other assignments $\mathbf{d}^{-tk}$:
\begin{equation}
  p(\dtk = j \mid \mathbf{d}^{-tk}) \propto
  \begin{cases}
    n_j^{-tk} & j \in \{1,\ldots,M_t\}, \\
    \alpha_0  & j = 0,
  \end{cases}
  \label{eq:crp}
\end{equation}
where $n_j^{-tk}$ is the accumulated association count assigned to landmark~$\lj$ excluding $z_{tk}$.
The CRP exposes two properties that the BPDA-GMM exploits. The per-landmark counts $n_j$ serve as sufficient statistics, and are updated incrementally in $O(1)$ per observation once the E-step weights are available (Sec.~\ref{sec:estep}).
The ``new-landmark'' branch also yields a scene-calibrated null hypothesis through $\alpha_0$
with a data-driven probability that adapts to scene density. To reduce excessive landmark birth in dense maps, we apply an exponential decay to recover the standard CRP at $\lambda = 0$:
\begin{equation}
  \alpha_0(t) = \alpha_0 \exp \bigl(-\lambda\,|\mathcal{M}_t|\bigr).
  \label{eq:alpha_decay}
\end{equation}

\subsection{Semantic-Constrained Likelihood}
\label{sec:likelihood}

We factorize the per-measurement likelihood under $\dtk = j$ into a geometric term and a semantic term:
\begin{equation}
\small
  p(\ztk \mid \xt,\lj,\dtk{=}j)
  = \underbrace{\mathcal{N}\!\bigl(z_{tk}^p;h(\xt,\lj^p),\,
    \bGamma\bigr)}_{\text{geometric}}\
    \underbrace{p(z_{tk}^s \mid \lj^s)}_{\text{semantic}},
  \label{eq:likelihood}
\end{equation}
where $h(\cdot)$ predicts landmark centroid in the camera frame, $\bGamma = \sigma_g^2 \mathbf{I}_3$ is the geometric noise covariance, and $p(z_{tk}^s \mid \lj^s)$ is the row of a class-conditional confusion matrix learned offline. 
Within Eq.~\eqref{eq:likelihood}, the confusion matrix acts as a soft score that down-weights landmarks whose accumulated class evidence diverges from the current detection.

To reduce computation, we restrict the candidate set of each measurement to landmarks of matching MAP class that also pass a Mahalanobis test on the marginal innovation covariance:
\begin{equation}
  \mathcal{H}_{tk} = \Big\{\,j \,\Big|\,
    \hat{\lj}^{\,s} = c_{tk}^*,\;\;
    \bm\nu_j^{\top}\bigl(\hat{\Sig}_j + \bGamma\bigr)^{-1}\bm\nu_j
    \le \chi^2_{3,0.99}\Big\},
  \label{eq:gate}
\end{equation}
where $c_{tk}^* = \arg\max_c z_{tk}^s(c)$ is the detection's MAP class, and $\bm\nu_j = z_{tk}^p - h(\hat{\xt}, \hat{\lj}^p)$ is the innovation.
Using the marginalized covariance, $\hat{\Sig}_j + \bGamma$,
keeps high-uncertainty, newly initialized landmarks inside the gate during early observations and reduces the candidate set from $M$ to $M_c = |\mathcal{H}_{tk}| \ll M$ in scenes with multiple classes. 

\begin{figure}
    \centering
    \includegraphics[width=0.98\linewidth]{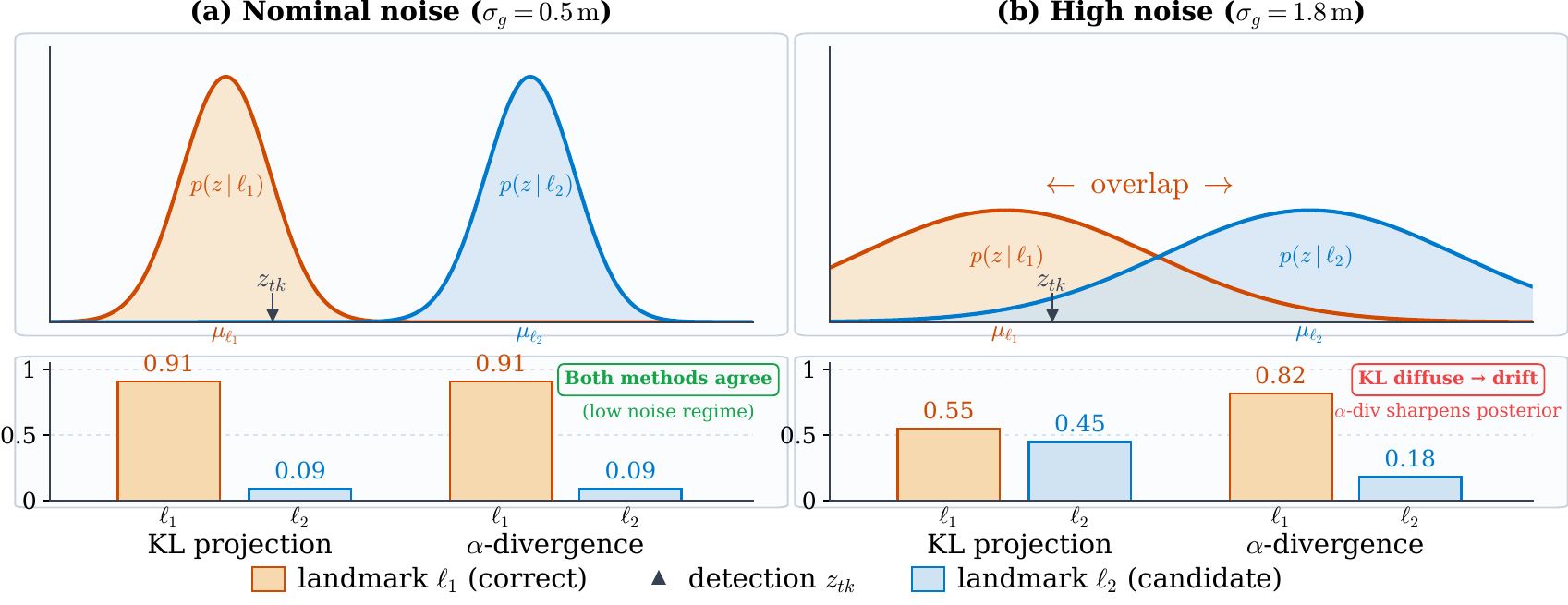}
    \caption{Effect of ambiguity-triggered posterior tempering. Under nominal noise, the untempered and tempered association weights agree. Under high geometric noise, the untempered posterior becomes diffuse, while tempering sharpens the dominant association and reduces drift toward competing candidates.
    }
    \label{fig:e2mschematic}
\end{figure}

\subsection{Bayesian Association Weight (E-step)}
\label{sec:estep}

Given current estimates $\hat{\X}^{(i)}, \hat{\LL}^{(i)}$, the posterior association weight combines the CRP prior~\eqref{eq:crp} with the likelihood~\eqref{eq:likelihood}:
\begin{equation}
\small
  w_{tkj}^{(i)} =
  \frac{n_j^{-tk}\,p(\ztk \mid \hat{\xt}^{(i)}, \lj^{(i)})}
       {Z_{tk}^{(i)}},
  w_{tk,0}^{(i)} =
  \frac{\alpha_0(t)\,p_0(\ztk)}{Z_{tk}^{(i)}},
  \label{eq:weight}
\end{equation}
where $Z_{tk}^{(i)} = \sum_{j' \in \mathcal{H}_{tk}}
n_{j'}^{-tk}\,p(\ztk \mid \hat{\xt}^{(i)}, \lj'^{(i)})
+ \alpha_0(t)\,p_0(\ztk)$ is the partition function, and $p_0(\ztk) = \mathcal{N}(z_{tk}^p;\,\mathbf{0},\,\sigma_0^2\mathbf{I})$ is a broad null-hypothesis prior with $\sigma_0 \gg \sigma_g$. The null-hypothesis weight is $w_{tk,\text{new}}^{(i)} = 1 - \sum_{j} \w^{(i)}$. Numerator and denominator span many orders of magnitude in dense maps, so we evaluate $Z_{tk}^{(i)}$ via the log-sum-exp identity:
\begin{equation}
  \log\!\sum_{j} e^{a_j} = a^{\star} + \log\!\sum_{j} e^{a_j - a^{\star}},
  \quad a^{\star} = \max_j a_j,
  \label{eq:lse}
\end{equation}
with $a_j = \log n_j^{-tk} + \log p(\ztk \mid \hat{\xt}^{(i)}, \lj^{(i)})$
for $j \in \mathcal{H}_{tk}$ and $a_0 = \log \alpha_0(t) + \log p_0(\ztk)$.
Under elevated geometric noise $\sigma_g$, likelihood ratios among competing in-gate candidates may shrink, producing a diffuse posterior. The resulting soft weights can pull the landmark update toward a weighted average of competing candidates, which may degrade trajectory estimates (Fig.~\ref{fig:e2mschematic}).
Therefore, we apply posterior tempering only when untempered weights are ambiguous:
\begin{equation}
  \underbrace{\max_j w_{tkj} < \tau_{\text{amb}}}_{\text{no dominant candidate}}
  \;\wedge\;
  \underbrace{w_{tk}^{(2)} / w_{tk}^{(1)} \geq \rho}_{\text{strong runner-up}},
  \label{eq:ambguard}
\end{equation}
where $w_{tk}^{(1)}, w_{tk}^{(2)}$ are the largest and second-largest weights.
If~\eqref{eq:ambguard} holds, we implement this robustness step as a temperature-scaled posterior, motivated by $\alpha$-divergence projection~\cite{ghalamkari2026em}. The weights are replaced by:
\begin{equation}
  \widetilde{w}_{tkj}^{(i)} \;\propto\;
  \bigl(n_j^{-tk}\,
  p(\ztk \mid \hat{\xt}^{(i)}, \lj^{(i)})\bigr)^{1/\alpha},
  \quad \alpha \in (0, 1].
  \label{eq:tempered}
\end{equation}
followed by normalization over $\mathcal{H}_{tk}\cup\{0\}$. Thus, $\alpha=1$ recovers the original posterior in~\eqref{eq:weight}, while smaller $\alpha$ increases the inverse temperature and sharpens ambiguous posteriors.

\subsection{Semantic Gaussian Landmark Update}
\label{sec:landmark}

Each landmark $\lj$ is represented as a semantic Gaussian $(\hat{\muu}_j, \hat{\Sig}_j, \hat{\ell}_j^s)$. 
Let $\bar{z}_j = \bigl(\sum_k w_{tkj}^{(i)}\, z_{tk}^{w}\bigr) / \sum_k w_{tkj}^{(i)}$ denote the weighted mean of back-projected measurements in the world frame ($z_{tk}^{w}$),
and $w_j^{\Sigma} = \sum_k w_{tkj}^{(i)}$ the total weight on landmark $j$. The innovation covariance and gain are then computed as:
\begin{equation}
  \mathbf{S}_j = \hat{\Sig}_j + \bGamma / w_j^{\Sigma},
  \qquad
  \mathbf{K}_j = \hat{\Sig}_j \mathbf{S}_j^{-1},
  \label{eq:innov}
\end{equation}
and the landmark is updated in closed form as:
\begin{align}
  \hat{\bm\mu}_j &\leftarrow \hat{\bm\mu}_j + \mathbf{K}_j(\bar{z}_j - \hat{\bm\mu}_j),
  \label{eq:kalman_mu}\\
  \hat{\Sig}_j &\leftarrow
    (\mathbf{I}-\mathbf{K}_j)\hat{\Sig}_j(\mathbf{I}-\mathbf{K}_j)^{\!\top}
    + \mathbf{K}_j\,\tfrac{\bGamma}{w_j^{\Sigma}}\,\mathbf{K}_j^{\!\top}.
  \label{eq:kalman_sigma}
\end{align}

The CRP count is a sufficient statistic and is updated by $O(1)$ increments per nonzero association weight after the E-step weights are computed. For a keyframe, the accumulated update is
\begin{equation}
n_j \leftarrow n_j + \sum_k w_{tkj}^{(i)},
\label{eq:crp_update}
\end{equation}



A new landmark is initialized whenever the null-hypothesis weights $w_{tk,0}^{(i)}$ exceeds a threshold $\theta_{\text{new}}$, with $\hat{\mu}_j^{(0)}$ set to the back-projected measurement and $\hat{\Sig}_j^{(0)} = s_0^2 \mathbf{I}$ for $s_0 = \max(6\sigma_g, 2.0)\,\text{m}$.
Object-level centroid extractions carry irreducible noise from viewpoint-dependent mask shape, so we floor the per-axis covariance at $\hat{\Sig}_j(d,d) \ge \sigma_g^2$ to prevent the gate from closing prematurely on re-observation. Landmarks with $n_j < \varepsilon$ are pruned every $P$ keyframes; their counts are retained for an additional $K_{\text{stale}}$ keyframes to preserve loop-closure identity~(Sec.~\ref{sec:loop_closure}).


Landmark class labels are not fixed at initialization but refined as evidence accumulates. Each landmark maintains a per-class vote accumulator $v_j(c)$, updated at every keyframe:
\begin{equation}
  v_j(c) \leftarrow v_j(c)
  + \sum_k w_{tkj}^{(i)}\, p(z_{tk}^s \mid \ell_j^s = c),
  \label{eq:class}
\end{equation}
where the MAP class estimate $\hat{\ell}_j^s = \arg\max_c v_j(c)$ allows early misclassifications to self-correct.
When two same-class landmarks are initialized for one physical object,
we merge pairs within Euclidean distance $d_{\text{merge}}$:
\begin{equation}
  \hat{\muu}_a \leftarrow \frac{n_a\hat{\muu}_a + n_b\hat{\muu}_b}{n_a+n_b},
  \qquad n_a \leftarrow n_a + n_b.
  \label{eq:merge}
\end{equation}
with the surviving identity inheriting the higher CRP count and the weighted-average position.

\subsection{Decoupled Pose--Landmark Optimization (M-step)}
\label{sec:mstep}

The M-step decouples trajectory estimation from semantic landmark optimization.  Object-level centroid extractors carry noise, an order of magnitude larger than per-keyframe Visual-Inertial Odometry (VIO) drift. 
We constrain the trajectory and landmarks through two disjoint factor groups.

\paragraph{Trajectory factors}  The trajectory $\hat{\X}$ is determined entirely by odometry through two factors per keyframe.
\begin{enumerate}
  \item A \textbf{between-factor} encodes the relative transform $\mathbf{u}_t$ with covariance $\Sig_t$ from the VIO source:
\begin{equation}
  f_{\text{o}}(\xt[t-1], \xt) =
  \bigl\| (\xt \ominus \xt[t-1]) \ominus \mathbf{u}_t \bigr\|^2_{\Sig_t^{-1}}.
  \label{eq:odom}
\end{equation}
  \item A \textbf{unary prior} anchors $\xt$ to its odometry-predicted value
  $\hat{\mathbf{x}}_t^{\text{odom}} = \hat{\mathbf{x}}_{t-1} \oplus \mathbf{u}_t$
with tight noise $\Sig_{\text{anchor}}$,
\begin{equation}
  f_{\text{a}}(\mathbf{x}_t) = \bigl\| \mathbf{x}_t \ominus
  \hat{\mathbf{x}}_t^{\text{odom}} \bigr\|^2_{\Sig_{\text{anchor}}^{-1}}.
  \label{eq:anchor}
\end{equation}
\end{enumerate}

\paragraph{Landmark factor}  Under Assumption~\ref{assump:for}, the per-measurement marginal likelihood under unknown association is a Gaussian mixture over candidate landmark:
\begin{equation}
  p(\ztk \mid \xt, \LL_{\mathcal{H}_{tk}})
  = \!\!\!\!\sum_{j \in \mathcal{H}_{tk} \cup \{0\}}\!\!\!\!
    p(\dtk = j \mid \mathbf{d}^{-tk})\,
    p(\ztk \mid \xt, \lj),
  \label{eq:gmm}
\end{equation}
where each component is a per-landmark Gaussian~\eqref{eq:likelihood} weighted by its CRP prior~\eqref{eq:crp}. Committing the full sum as a factor is incompatible with iSAM2's least-squares linearization. We replace the sum with a max to obtain the \emph{max-mixture} factor that retains the dominant component: 
\begin{equation}
\small
  f_{tk}(\xt, \LL_{\mathcal{H}'_{tk}}) =
  \max_{j \in \mathcal{H}'_{tk}}
  \; p(\ztk \mid \xt, \lj)\,
     p(\dtk = j \mid \mathbf{d}^{-tk}),
  \label{eq:mm}
\end{equation}
restricted to mature landmarks $\mathcal{H}'_{tk} = \{j \in \mathcal{H}_{tk} : n_j \ge 2\}$. The max-mixture is exact when one component dominates, introducing a Gaussian residual after linearization and preserving compatibility with the nonlinear least-squares back-end.

Inspired by SLIDE-SLAM~\cite{liu2025slideslam}, we override the pose Jacobian $\partial f_{tk}/\partial x_t$ to zero, so that during iSAM2 linearization, the factor produces no gradient with respect to the robot pose. Only the landmark Jacobian active mixture component, evaluated at the max-attaining $\partial f_{tk} / \partial \lj^p = -\mathbf{L}^{-1} \mathbf{R}_{cw}$ (where $\mathbf{L}$ is the Cholesky factor of $\bGamma$ and $\mathbf{R}_{cw}$ the camera-to-world rotation) contributes to the update.
The maturity gate $n_j \ge 2$ ensures that high-uncertainty newly initialized landmarks are refined solely through the Kalman layer~\eqref{eq:kalman_mu}--\eqref{eq:kalman_sigma} before they begin influencing the factor graph.
\begin{algorithm}[t]
\caption{BPDA-GMM: per-keyframe update}
\label{alg:bpda}
\SetAlgoLined\DontPrintSemicolon
\KwIn{$\mathbf{u}_t$,  $\Sig_t$, $\Z_t$, 
      $\hat{\X}, \hat{\LL}, \{n_j\}, \{v_j\}$}
\KwOut{Updated $\hat{\X},\hat{\LL},\{n_j\},\{v_j\}$}
Predict $\hat{\mathbf{x}}_t^{\text{odom}} \leftarrow \hat{\mathbf{x}}_{t-1} \oplus \mathbf{u}_t$\;
Add between-factor~\eqref{eq:odom} and  anchor prior~\eqref{eq:anchor} to graph\;
\For{$i=1$ \KwTo $I$}{
  \For{each $\ztk \in \Z_t$}{
    Build $\mathcal{H}_{tk}$ via~\eqref{eq:gate}\;
    Compute $w_{tkj}^{(i)}$ via~\eqref{eq:weight} in log domain~\eqref{eq:lse}\;
    \If{\textup{ambiguity gate~\eqref{eq:ambguard} fires}}{
      Replace $\w^{(i)}$ with tempered weights~\eqref{eq:tempered}\;
    }
  }
  \If{$i=1$}{
    Init landmarks with $w_{tk,\text{new}}\ge\theta_{\text{new}}$\;
    Update counts~\eqref{eq:crp_update} and class votes~\eqref{eq:class}\;
    Merge same-class pairs within $d_{\text{merge}}$ via~\eqref{eq:merge}\;
    \textbf{Loop closure:} match $\mathbf{d}_t$ against history; for
    each hit, apply~\eqref{eq:merge} on pairs and add~\eqref{eq:lc}\;
  }
  Update $(\hat{\muu}_j,\hat{\Sig}_j)$ via
  \eqref{eq:kalman_mu}--\eqref{eq:kalman_sigma}\;
}
\ForEach{$\ztk$ with $w_{tk,0}^{(I)} < \theta_{\text{factor}}$ and
        $\mathcal{H}'=\{j\in\mathcal{H}_{tk}:n_j\ge 2\}\ne\emptyset$}{
  Add max-mixture factor~\eqref{eq:mm} with
  $\partial f/\partial\xt=\mathbf{0}$\;
}
$\hat{\X},\hat{\LL}\leftarrow$ iSAM2.\texttt{update()}\;
Sync $\hat{\muu}_j$ from iSAM2 back to landmark map\;
\If{$t \bmod P = 0$}{
 Prune landmarks with $n_j<\varepsilon$\;
 Discard retained counts older than $K_{\text{stale}}$\;}
\end{algorithm}

\subsection{Semantic Loop Closure via CRP Re-identification}
\label{sec:loop_closure}

When the robot revisits a region, the same-class gate $\mathcal{H}_{tk}$ already returns the dormant landmark as a candidate, and its CRP count $n_j$, retained through staleness pruning for up to $K_{\text{stale}}$ keyframes, biases the weight~\eqref{eq:weight} toward the old identity instead of a fresh null hypothesis.
Each keyframe stores a compact semantic place descriptor:
\begin{equation}
  \mathbf{d}_t = \bigl(h_t,\;
    \{(\hat{\ell}^{\,s}_j,\hat{\bm{\mu}}_j - \hat{\mathbf{x}}_t)\}_{j\in\mathcal{V}_t}\bigr),
  \label{eq:place}
\end{equation}
where $\mathcal{V}_t$ is the set of in-range landmarks and $h_t$ a class-histogram hash.
Keyframes with matching hashes are verified using a per-class Wasserstein distance between radial relative-position sets. If the distance is below $\tau_w$, the resulting landmark pairs $\mathcal{P}_{tt'}\subset\mathcal{V}_t\times\mathcal{V}_{t'}$ are merged using map-side closure via~\eqref{eq:merge}, and a pose-side loop-closure factor is added:

\begin{equation}
  f_{\text{lc}}(\mathbf{x}_t, \mathbf{x}_{t'}) =
  \bigl\| \mathbf{x}_{t'} \ominus (\mathbf{x}_t \oplus \hat{\mathbf{u}}_{tt'})
  \bigr\|^{2}_{\bm{\Sigma}_{\text{lc}}^{-1}},
  \label{eq:lc}
\end{equation}
where $\hat{\mathbf{u}}_{tt'} \in \mathfrak{se}(3)$ is the rigid transform that least-squares aligns the matched pairs in $\mathcal{P}_{tt'}$,
and $\bm{\Sigma}_{\text{lc}}$ is scaled by alignment residual.
In addition, if a closure incorrectly matches distinct landmark sets, the CRP weights of the reused identities collapse on the next keyframe, and iSAM2 retracts the closure factor through standard relinearization and outlier rejection.
\section{Experimental Evaluation}
\label{sec:experiments}
\subsection{Implementation and Baselines}  

BPDA-GMM is implemented in C++ with ROS Noetic, GTSAM~4.2~\cite{dellaert2012factor}, and OpenCV~4.2. Object centroids are triangulated from stereo detections, and TensorRT FP16 is used for detector inference on a Jetson Xavier AGX. All baselines share the same odometry, measurement model, landmark update pipeline, and iSAM2 back-end~\cite{kaess2012isam2}; only the association policy is changed.
We evaluate pose accuracy using absolute trajectory error (ATE $\downarrow$) and relative pose error (RPE $\downarrow$), map quality using landmark Chamfer distance $\downarrow$ and geometric uncertainty $\downarrow$, and semantic quality using class-belief entropy $\downarrow$ and semantic accuracy $\uparrow$.

\begin{table}[!ht]
  \caption{Simulation scenarios used in our benchmark.}
  \label{tab:sim_scenarios}
  \centering
  \scriptsize
    \begin{tabularx}{0.49\textwidth}{
    >{\raggedright\arraybackslash}p{0.088\textwidth} 
    >{\centering\arraybackslash}p{0.104\textwidth} 
    >{\centering\arraybackslash}p{0.087\textwidth}  
    >{\centering\arraybackslash}p{0.035\textwidth}
    >{\centering\arraybackslash}X
    }
    \toprule
    Scenario
    & Landmark model & Extent (m) & $\#$classes & Obs. range \\
    \midrule

    Figure Eight &
    30 random point &
    $8 \times 6$&
    6 &
    6\,m \\

    Outdoor Loop &
    9 sparse semantic &
    $\approx 80 \times 80$ &
    3 &
    15\,m \\

    Complex Urban &
    repeated urban &
    $\approx 190 \times 155$ &
    6 &
    22\,m \\

    Victoria Park &
    151 Ground truth&
    $\approx 280 \times 265$ &
    5 &
    25\,m \\

    City10000 &
    route-conditioned &
    $\approx 105 \times 105$ &
    6 &
    14\,m \\
    \bottomrule
  \end{tabularx}
\end{table}

\begin{table}[t]
\centering
\caption{Simulation results across all benchmark scenarios. \small{Best values are shown in bold and second-best values are underlined.}}
\label{tab:sim_results_nonoracle}
\scriptsize
\begin{tabularx}{0.49\textwidth}{
>{\raggedright\arraybackslash}p{0.037\textwidth} 
>{\raggedright\arraybackslash}p{0.088\textwidth} 
>{\centering\arraybackslash}X 
>{\centering\arraybackslash}X 
>{\centering\arraybackslash}X 
>{\centering\arraybackslash}X 
>{\centering\arraybackslash}X 
>{\centering\arraybackslash}X 
}
\toprule
Scenario & Method & ATE$\downarrow$ & RPE$\downarrow$ & lm$_\mathrm{cham}$$\downarrow$ & geo$_\mathrm{unc}$$\downarrow$ & sem$_\mathrm{ent}$$\downarrow$ & sem$_\mathrm{acc}$$\uparrow$ \\
\midrule

& MHJCBB\cite{wang2018robust} & 0.1737 & 0.0685 & 0.2861 & \underline{0.0016} & 0.8439 & \textbf{0.9385} \\
& MH-iSAM2\cite{hsiao2019mh} & 0.1740 & 0.0686 & 0.2852 & \textbf{0.0015} & 0.8443 & \textbf{0.9385} \\
& Gauss PDA\cite{bowman2017probabilistic} & 0.1731 & \underline{0.0680} & 0.2750 & \textbf{0.0015} & 0.8459 & \textbf{0.9385} \\
Figure
& MMSS\cite{doherty2019multimodal} & 0.1683 & 0.0684 & \underline{0.2740} & \underline{0.0016} & 0.8350 & \underline{0.9242} \\
Eight
& SGBA\cite{ji2024sgba} & \underline{0.1611} & 0.0691 & 0.2806 & 0.0018 & \underline{0.8287} & 0.9048 \\
& k-best\cite{michael2022probabilistic} & 0.1666 & 0.0686 & 0.2742 & 0.0017 & 0.8369 & 0.9077 \\
& \textbf{BPDA-GMM} & \textbf{0.1438} & \textbf{0.0672} & \textbf{0.2154} & 0.0028 & \textbf{0.8040} & 0.9020 \\
\midrule

& MHJCBB\cite{wang2018robust} & 44.1541 & 0.8220 & 61.9860 & 0.6615 & 0.4572 & 0.0000 \\
& MH-iSAM2\cite{hsiao2019mh} & 54.1171 & 0.8482 & 69.9649 & \underline{0.6615} & 0.4572 & 0.0000 \\
Outdoor & Gauss PDA\cite{bowman2017probabilistic} & \underline{31.1061} & 0.9011 & \underline{53.9796} & \textbf{0.6554} & 0.4572 & \underline{0.3333} \\
Loop & MMSS\cite{doherty2019multimodal} & 51.7176 & 0.8253 & 68.6181 & 0.7466 & 0.4581 & 0.1111 \\
& SGBA\cite{ji2024sgba} & 33.0799 & \underline{0.7836} & 54.8024 & 0.7478 & \textbf{0.4519} & \underline{0.3333} \\
& k-best\cite{michael2022probabilistic} & 54.6456 & 0.8335 & 68.5899 & 0.7466 & \underline{0.4570} & \underline{0.3333} \\
& \textbf{BPDA-GMM} & \textbf{14.2324} & \textbf{0.6590} & \textbf{32.5817} & 0.7678 & 0.4597 & \textbf{0.8889} \\
\midrule

& MHJCBB\cite{wang2018robust} & 4.6379 & 0.3498 & 12.4009 & 0.0785 & 0.9408 & 0.6738 \\
& MH-iSAM2\cite{hsiao2019mh} & 4.6380 & 0.3498 & 12.4009 & 0.0785 & 0.9408 & 0.6738 \\
Complex & Gauss PDA\cite{bowman2017probabilistic} & \underline{4.5801} & 0.3496 & 12.3700 & 0.0781 & 0.9415 & \underline{0.6968} \\
Urban & MMSS\cite{doherty2019multimodal} & 4.7962 & \underline{0.3493} & 12.3977 & 0.0781 & 0.9390 & 0.6862 \\
& SGBA\cite{ji2024sgba} & 4.8357 & 0.3497 & \underline{12.3525} & \textbf{0.0781} & \underline{0.9384} & 0.6935 \\
& k-best\cite{michael2022probabilistic} & 4.8147 & 0.3496 & 12.3834 & \underline{0.0781} & 0.9390 & 0.6738 \\
& \textbf{BPDA-GMM} & \textbf{4.5237} & \textbf{0.3492} & \textbf{12.2059} & 0.0836 & \textbf{0.9343} & \textbf{0.7112} \\
\midrule

& MHJCBB\cite{wang2018robust} & 3.9287 & 1.0778 & 3.7652 & 0.0762 & 0.7696 & 0.4628 \\
& MH-iSAM2\cite{hsiao2019mh} & 3.4051 & 1.0832 & 2.9906 & 0.0767 & 0.7714 & 0.4357 \\
City& Gauss PDA\cite{bowman2017probabilistic} & 7.9954 & 1.1225 & 5.1991 & 0.0776 & 0.7704 & 0.2532 \\
10000& MMSS\cite{doherty2019multimodal} & 7.0389 & 1.2684 & 5.2275 & 0.0802 & 0.7721 & 0.2815 \\
& SGBA\cite{ji2024sgba} & \underline{3.0256} & \textbf{0.9636} & \textbf{2.3267} & \underline{0.0761} & \underline{0.7690} & \textbf{0.6234} \\
& k-best\cite{michael2022probabilistic} & \textbf{2.9864} & 0.9687 & 2.3348 & \textbf{0.0759} & \textbf{0.7659} & \underline{0.6174} \\
& \textbf{BPDA-GMM} & 3.0326 & \underline{0.9637} & \underline{2.3304} & \underline{0.0761} & \underline{0.7690} & \textbf{0.6234} \\
\midrule

& MHJCBB\cite{wang2018robust} & 14.8020 & 0.7169 & \underline{18.4779} & \underline{52.0789} & \textbf{1.4204} & 0.6286 \\
& MH-iSAM2\cite{hsiao2019mh} & 10.9810 & 0.6997 & 19.0820 & 199.4214 & 1.5386 & \underline{0.7429} \\
Victoria & Gauss PDA\cite{bowman2017probabilistic} & \underline{10.5024} & \underline{0.6869} & 18.8324 & 52.1332 & \underline{1.4205} & \underline{0.7429} \\
Park& MMSS\cite{doherty2019multimodal} & 11.0187 & 0.7010 & 19.0962 & 199.4035 & 1.5387 & \underline{0.7429} \\
& SGBA\cite{ji2024sgba} & 10.6185 & 0.6910 & 18.8944 & 222.1721 & 1.5461 & \underline{0.7429} \\
& k-best\cite{michael2022probabilistic} & 15.0317 & 0.7241 & 18.5680 & 222.1337 & 1.5464 & 0.6571 \\
& \textbf{BPDA-GMM} & \textbf{2.6967} & \textbf{0.4000} & \textbf{13.5608} & \textbf{33.2951} & 1.4774 & \textbf{0.8286} \\
\bottomrule
\end{tabularx}
\end{table}

\begin{table}[t]
  \caption{Quantitative results of object-level representation. 
  }
  \label{tab:real_results}
    \scriptsize
    \begin{tabularx}{0.49\textwidth}{
    >{\raggedright\arraybackslash}p{0.09\textwidth} 
    >{\raggedright\arraybackslash}p{0.038\textwidth} 
    >{\centering\arraybackslash}p{0.021\textwidth}
    >{\centering\arraybackslash}p{0.05\textwidth}
    >{\centering\arraybackslash}p{0.05\textwidth}
    >{\centering\arraybackslash}p{0.035\textwidth}
    >{\centering\arraybackslash}X 
    }
    \toprule
    Method & Semantic & G.T. & Estimation & Precision$\uparrow$ & Recall$\uparrow$ & F1$\uparrow$ \\
    \midrule
    \multirow{6}{*}{Gauss PDA\cite{bowman2017probabilistic}}
    & chair    & 36 & 38 & 0.737 & 0.778 & 0.757  \\
    & table    & 21 & 1  & 1.000 & 0.048 & 0.091 \\
    & monitor  & 18 & 13 & 0.615 & 0.444 & 0.516 \\
    & fridge   & 5  & 16 & 0.312 & 1.000 & 0.476 \\
    & plant    & 4  & 2  & 1.000 & 0.500 & 0.667 \\
    & overall  & 84 & 70 & 0.629 & 0.524 & 0.571 \\
    \midrule

    \multirow{6}{*}{MMSS\cite{doherty2019multimodal}}
    & chair    & 36 & 31 & 0.806 & 0.694 & 0.746 \\
    & table    & 21 & 1  & 1.000 & 0.048 & 0.091 \\
    & monitor  & 18 & 10 & 0.700 & 0.389 & 0.500 \\
    & fridge   & 5  & 7  &  0.429 & 0.600 & 0.500 \\
    & plant    & 4  & 1  & 1.000 & 0.250 & 0.400 \\
    & overall  & 84 & 50 & 0.740 & 0.440 & 0.552 \\
    \midrule

    \multirow{6}{*}{MMNH \cite{doherty2020probabilistic}}
    & chair    & 36 & 29 & 0.793 & 0.639 & 0.708 \\
    & table    & 21 & 2  & 1.000 & 0.095 & 0.174 \\
    & monitor  & 18 & 10 & 0.700 & 0.389 & 0.500 \\
    & fridge   & 5  & 9  & 0.333 & 0.600 & 0.429 \\
    & plant    & 4  & 1  & 1.000 & 0.250 & 0.400 \\
    & overall  & 84 & 51 & 0.706 & 0.429 & 0.533 \\
    \midrule

    \multirow{6}{*}{MHJCBB\cite{wang2018robust}}
    & chair   & 36 & 56 & 0.554 & 0.861 & 0.674 \\
    & table   & 21 & 5  & 1.000 & 0.238 & 0.385 \\
    & monitor & 18 & 13 & 0.615 & 0.444 & 0.516 \\
    & fridge  & 5  & 9  & 0.333 & 0.600 & 0.429 \\
    & plant   & 4  & 3  & 1.000 & 0.750 & 0.857 \\
    & overall & 84 & 86 & 0.581 & 0.595 & 0.588 \\
    \midrule

    \multirow{6}{*}{MH-iSAM2\cite{hsiao2019mh}}
    & chair   & 36 & 59  & 0.576 & 0.944 & 0.716 \\
    & table   & 21 & 9   & 0.778 & 0.333 & 0.467 \\
    & monitor & 18 & 20  & 0.700 & 0.778 & 0.737 \\
    & fridge  & 5  & 7   & 0.571 & 0.800 & 0.667 \\
    & plant   & 4  & 4   & 1.000 & 1.000 & 1.000 \\
    & overall & 84 & 101 & 0.636 & 0.750 & 0.689 \\
    \midrule

    \multirow{6}{*}{k-best\cite{michael2022probabilistic}}
    & chair   & 36 & 53  & 0.604 & 0.889 & 0.719 \\
    & table   & 21 &  9  & 0.889 & 0.381 & 0.533 \\
    & monitor & 18 & 22  & 0.591 & 0.722 & 0.650 \\
    & fridge  & 5  & 17  & 0.235 & 0.800 & 0.364 \\
    & plant   & 4  & 3   & 1.000 & 0.750 & 0.857 \\
    & overall & 84 & 104 & 0.577 & 0.714 & 0.638 \\
    \midrule

    \multirow{6}{*}{SGBA\cite{ji2024sgba}}
    & chair   & 36 & 35 & 0.714 & 0.694 & 0.704 \\
    & table   & 21 & 14 & 0.929 & 0.619 & 0.743 \\
    & monitor & 18 & 19 & 0.632 & 0.667 & 0.649 \\
    & fridge  & 5  & 18 & 0.278 & 1.000 & 0.435 \\
    & plant   & 4  & 4  & 0.750 & 0.750 & 0.750 \\
    & overall & 84 & 86 & 0.644 & 0.690 & 0.667\\
    \midrule

    \multirow{6}{*}{SlideSLAM\cite{liu2025slideslam}}
    & chair   & 36 & 38 & 0.684 & 0.722 & 0.703 \\
    & table   & 21 & 6  & 0.667 & 0.190 & 0.296 \\
    & monitor & 18 & 3  & 0.667 & 0.111 & 0.190 \\
    & fridge  & 5  & 0  & 0.000 & 0.000 & 0.000\\
    & plant   & 4  & 0  & 0.000 & 0.000 & 0.000 \\
    & overall & 84 & 47 & 0.681 & 0.381 & 0.489 \\
    \midrule

    \multirow{6}{*}{\textbf{BPDA-GMM}}
    & chair   & 36 & 35 & 0.771 & 0.750 & 0.761 \\
    & table   & 21 & 14 & 0.929 & 0.619 & 0.743 \\
    & monitor & 18 & 20 & 0.700 & 0.778 & 0.737 \\
    & fridge  & 5  & 7  & 0.571 & 0.800 & 0.667 \\
    & plant   & 4  & 3  & 1.000 & 0.750 & 0.857 \\
    & overall & 84 & 77 & \textbf{0.772} & \textbf{0.726} & \textbf{0.748} \\
    \bottomrule
\end{tabularx}
\end{table}

\begin{figure}[!ht]
  \centering
 \includegraphics[width=0.49\textwidth]{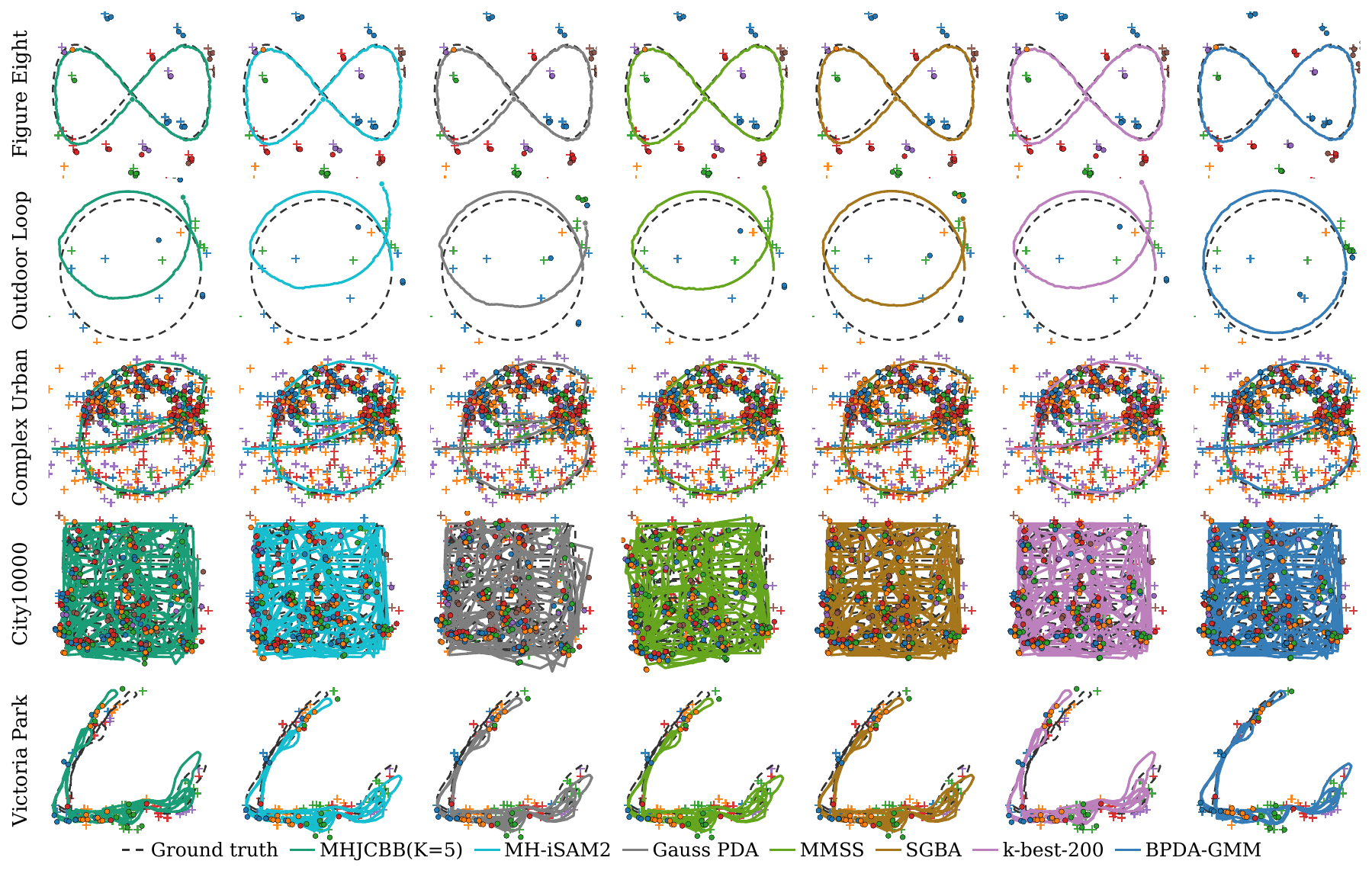}
  \caption{Estimated trajectories on all benchmark scenarios. BPDA-GMM remains close to the ground truth across dense, sparse, and perceptually aliased environments, while several baselines accumulate large drift in high-ambiguity cases.}
  \label{fig:sim_traj}
\end{figure}
\subsection{Simulation Benchmark}

We evaluate data-association robustness in five simulated scenarios, summarized in Table~\ref{tab:sim_scenarios}. The benchmark includes dense indoor trajectories, sparse outdoor loops, perceptually aliased urban layouts, and two dataset-driven routes from the MH-iSAM2~\cite{hsiao2019mh} benchmark. Detections are corrupted by Gaussian geometric noise and a class-confusion model, while odometry is perturbed by translational and rotational noise.

\begin{figure}[!ht]
  \centering
  \includegraphics[width=0.48\textwidth]{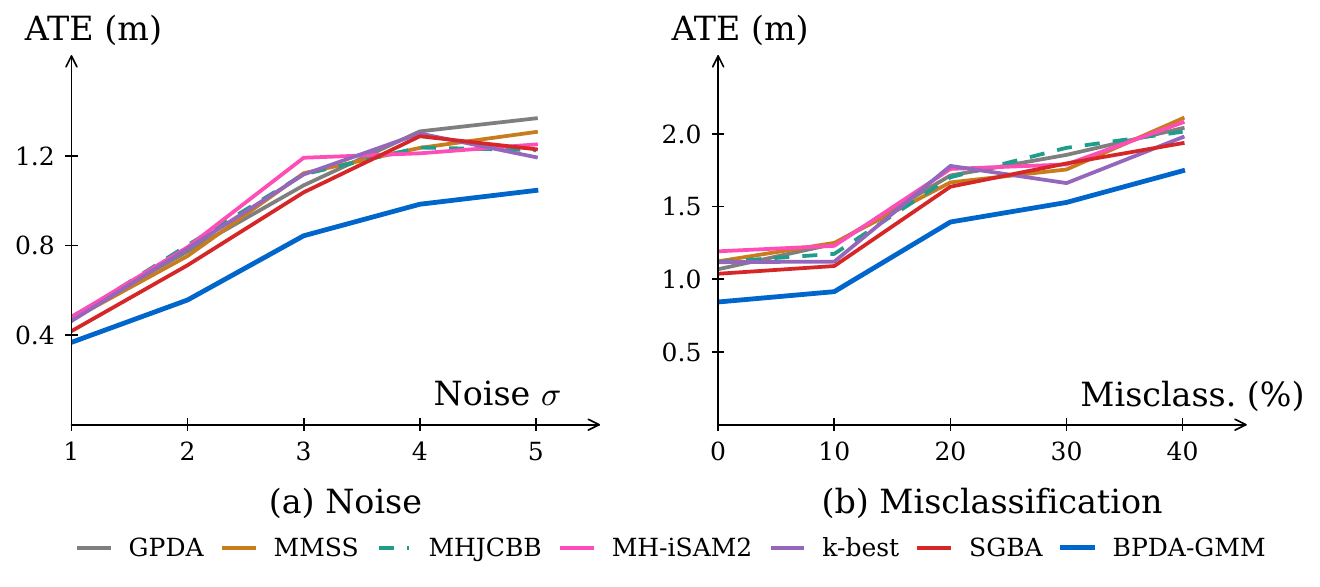}
  \caption{Robustness to odometry noise and semantic misclassification. BPDA-GMM shows slower ATE growth as noise and classifier errors increase, indicating improved resistance to ambiguous or incorrect associations.}
  \label{fig:noise_sweep}
\end{figure}


Table~\ref{tab:sim_results_nonoracle} reports the quantitative results, and Fig.~\ref{fig:sim_traj} shows representative trajectory overlays. BPDA-GMM achieves the best ATE in four of five scenarios and the best or tied-best RPE in four of five. The largest gains appear in high-ambiguity cases. On Outdoor Loop, BPDA-GMM reduces ATE from 31.11~m for the next-best PDA baseline to 14.23~m, while improving semantic accuracy to 0.889. On Victoria Park, ATE decreases from 10.50~m to 2.70~m, and landmark Chamfer distance is also reduced. In Complex Urban, BPDA-GMM obtains the best ATE, RPE, lm$_{\text{cham}}$, sem$_{\text{ent}}$, and sem$_{\text{acc}}$, indicating robustness to same-class perceptual aliasing. Fig.~\ref{fig:noise_sweep} further shows that BPDA-GMM degrades more slowly as odometry noise and detector misclassification increase.

\begin{figure*}[!ht]
    \centering
    \includegraphics[width=\textwidth]{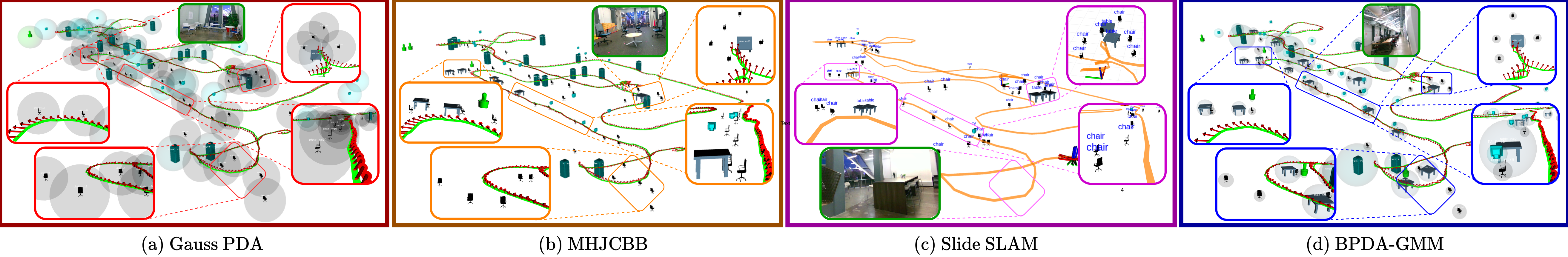}
    \caption{Qualitative object-map comparison on the real indoor sequence. BPDA-GMM produces a compact semantic map with fewer duplicate landmarks and lower uncertainty, while hard-association or fixed-hypothesis baselines suffer from missed objects or over-instantiation.}
    \label{fig:qualitative}
\end{figure*}

\subsection{Real Indoor Dataset}
\label{sec:indoor}
We also evaluate on an indoor sequence collected with a Falcon~250 quadrotor equipped with a RealSense D455 stereo camera. The scene contains 84 ground-truth objects from five classes: chair, table, monitor, fridge, and plant. Table~\ref{tab:real_results} reports object-level precision, recall, and F1 score. BPDA-GMM achieves the highest overall F1 score, 0.748, with the best precision, 0.772, and the second-best recall, 0.726. Multi-hypothesis baselines achieve high recall but over-instantiate landmarks; for example, MH-iSAM2 estimates 101 landmarks for 84 objects, reducing precision to 0.636. In contrast, BPDA-GMM estimates 77 landmarks, showing that the CRP prior suppresses duplicate landmark creation while preserving most true objects. BPDA-GMM is also the only method with nonzero F1 across all five classes, including small or weakly detected objects such as plants.

Fig.~\ref{fig:qualitative} compares estimated object maps across four representative methods. Landmarks are rendered as translucent spheres scaled by uncertainty. Gauss PDA produces large overlapping spheres and scatters multiple landmarks for the same physical object, reflecting an inability of soft PDA to commit when association is ambiguous. MHJCBB misclassifies fridges and strings duplicates along revisited corridors, since its hard top-$K$ assignment cannot recover from a wrong initial class. SlideSLAM, whose hard-new policy fixes class at landmark birth, fails to instantiate rare or weakly-detected classes (fridge, plant), leaving large regions of the map empty. BPDA-GMM produces a clean map with one compact, low-uncertainty landmark per physical object, demonstrating that the CRP prior and the soft semantic posterior together suppress both spurious duplicates and missing-class failures.



\subsection{Computational, Ablation, and Sensitivity Analysis}

\label{sec:timing}

\begin{table}[t]
  \caption{Per-keyframe processing time (mean/max, ms)}
  \scriptsize
  \centering
  \label{tab:timing}
    \begin{tabularx}{0.49\textwidth}{
    >{\raggedright\arraybackslash}p{0.093\textwidth} 
    >{\centering\arraybackslash}p{0.068\textwidth} 
    >{\centering\arraybackslash}p{0.068\textwidth}
    >{\centering\arraybackslash}p{0.068\textwidth}
    >{\centering\arraybackslash}X
    }
    \toprule
    Method & E-step & lm-update & M-step & Total \\
    \midrule
    Gauss PDA\cite{bowman2017probabilistic}& 0.0274/0.2212 & 0.0121/0.0848 & 1.1974/10.1492 & 1.2369/10.4552\\
    MMSS\cite{doherty2019multimodal}       & 0.0216/0.2170 & 0.0097/0.0540 & 1.1204/10.7125 & 1.1517/10.9835\\
    MMNH\cite{doherty2020probabilistic}    & 0.0222/0.2835 & 0.0119/0.4134 & 1.1763/9.5270  & 1.2104/10.2239\\
    MHJCBB\cite{wang2018robust}            & 0.0458/0.4714 & 0.0179/0.1810 & 1.2625/7.0591  & 1.3262/7.7115\\
    MH-iSAM2\cite{hsiao2019mh}        & 0.6668/2.8993 & 111.95/455.52 & 111.36/453.96  & 223.98/912.37\\
    k-best\cite{michael2022probabilistic}  & 0.0459/0.4588 & 0.0158/0.0935 & 1.1194/7.9667  & 1.1811/8.5190\\
    SGBA\cite{ji2024sgba}                  & 0.0522/0.4136 & 0.0139/0.5610 & 1.1198/16.3624 & 1.1859/17.337\\
    SlideSLAM\cite{liu2025slideslam}     & 0.0463/0.4315 & 0.0015/0.0126 & 1.0906/6.0363 & 1.1384/6.4804\\
    \textbf{BPDA-GMM} & \textbf{0.0290/0.1474} & \textbf{0.0087/0.0462} & \textbf{0.7681/3.7738} & \textbf{0.8058/3.9674}\\
    \bottomrule
  \end{tabularx}
\end{table}

Table~\ref{tab:timing} reports per-keyframe runtime on the Jetson Xavier AGX, and a desktop machine with an Intel Core~i7 is used only for RViz visualization. BPDA-GMM achieves the lowest mean and maximum runtime, $0.806$\,ms and $3.97$\,ms, respectively. Compared with soft-association baselines, it reduces mean runtime by about $30$--$50\%$, while MH-iSAM2 is over two orders of magnitude slower due to explicit multi-hypothesis optimization. BPDA-GMM therefore supports real-time embedded semantic SLAM and sustains $>\!10$\,Hz end-to-end.

\begin{table}[t]
  \caption{Ablation study when removing DP prior, semantic likelihood, soft Bayesian association, and decoupled back-end.}
  \scriptsize
  \centering
  \label{tab:abl}
    \begin{tabularx}{0.49\textwidth}{
    >{\raggedright\arraybackslash}p{0.19\textwidth} 
    >{\centering\arraybackslash}X 
    >{\centering\arraybackslash}X
    >{\centering\arraybackslash}X
    >{\centering\arraybackslash}X
    }
    \toprule
    Method & ATE$\downarrow$ & RPE$\downarrow$ & lm$_\mathrm{cham}$$\downarrow$ & sem$_\mathrm{acc}$$\uparrow$ \\
    \midrule
    Full BPDA-GMM  
    & 3.0326  & 0.9637 & 2.3304 & 0.6234\\
    w/o DP prior~\eqref{eq:dp}\eqref{eq:crp}\eqref{eq:alpha_decay} 
    & 7.9954  & 1.1225 & 5.1991 & 0.2532\\
    w/o Semantic likelihood~\eqref{eq:likelihood} 
    & 4.0925  & 1.1148 & 2.9201 & 0.5276\\
    w/o soft Bayesian association~\eqref{eq:weight}
    & 3.6070  & 1.1262 & 3.2970 & 0.5141\\
    w/o decoupling~\eqref{eq:mm}
    & 14.1801 & 2.6226 & 6.8444 & 0.1960\\
    \bottomrule
  \end{tabularx}
\end{table}

Table~\ref{tab:abl} evaluates the contribution of the main BPDA-GMM components. Removing the DP prior causes the largest semantic degradation, reducing semantic accuracy from $0.6234$ to $0.2532$, because the model loses its CRP-based new-landmark mechanism and reverts to a fixed-cardinality association behavior. Removing the semantic likelihood weakens class-consistent association and increases both trajectory and map errors. Replacing soft Bayesian association with a non-probabilistic alternative also degrades all metrics, showing the importance of maintaining association uncertainty. Removing the decoupled back-end causes the largest pose degradation, increasing ATE to $14.18$\,m. This confirms that zero-pose-Jacobian semantic factors are essential for preventing noisy object detections from corrupting the trajectory.

\begin{table}[t]
  \caption{Sample of sensitivity analysis. \small{Each setting is averaged over 42 random seeds while varying one hyperparameter at a time.}}
  \scriptsize
  \centering
  \label{tab:sensi}
  \begin{tabularx}{0.49\textwidth}{
    >{\raggedright\arraybackslash}p{0.1\textwidth}
    >{\centering\arraybackslash}X
    >{\centering\arraybackslash}X
    >{\centering\arraybackslash}X
    >{\centering\arraybackslash}X}
    \toprule
    Parameter & ATE$\downarrow$ & RPE$\downarrow$ & lm$_\mathrm{cham}\downarrow$ & sem$_\mathrm{acc}\uparrow$ \\
    \midrule
    $\alpha_0=0.10$ & 6.8849 & 1.1447 & 4.5275 & 0.3082 \\
    $\alpha_0=0.25$ & 6.5501 & 1.1324 & 4.3025 & 0.3207 \\
    $\alpha_0=0.50$ & 6.1767 & 1.1130 & 4.2393 & 0.3301 \\
    $\alpha_0=0.75$ & 6.9687 & 1.1345 & 4.5081 & 0.3013 \\
    $\alpha_0=1.00$ & 6.9771 & 1.1328 & 4.5748 & 0.3075 \\
    $\lambda=0.000$ & 6.5819 & 1.1345 & 4.3437 & 0.3228 \\
    $\lambda=0.001$ & 6.2834 & 1.1235 & 4.1783 & 0.3318 \\
    $\lambda=0.002$ & 6.4067 & 1.1397 & 4.2787 & 0.3263 \\
    $\lambda=0.004$ & 6.9694 & 1.1328 & 4.4953 & 0.3156 \\
    $\lambda=0.008$ & 7.0796 & 1.1406 & 4.5329 & 0.3100 \\
    no tempering & 6.9960 & 1.1387 & 4.5025 & 0.3036 \\
    $\alpha=0.15$ & 6.7640 & 1.1481 & 4.4374 & 0.3023 \\
    $\alpha=0.30$ & 6.2834 & 1.1235 & 4.1783 & 0.3318 \\
    $\alpha=0.60$ & 6.5184 & 1.1201 & 4.2986 & 0.3266 \\
    $\alpha=0.80$ & 6.8014 & 1.1362 & 4.4017 & 0.3126 \\
    $\tau_\mathrm{amb}=0.50$ & 6.9960 & 1.1387 & 4.5025 & 0.3036 \\
    $\tau_\mathrm{amb}=0.65$ & 6.7979 & 1.1409 & 4.4446 & 0.3078 \\
    $\tau_\mathrm{amb}=0.80$ & 6.2834 & 1.1235 & 4.1783 & 0.3318 \\
    $\tau_\mathrm{amb}=0.90$ & 6.3998 & 1.1256 & 4.2154 & 0.3224 \\
    $\tau_\mathrm{amb}=1.00$ & 6.3998 & 1.1256 & 4.2154 & 0.3224 \\
    $d_\mathrm{merge}=0.5$ & 6.5202 & 1.1239 & 4.3070 & 0.3115 \\
    $d_\mathrm{merge}=1.0$ & 6.7262 & 1.1049 & 4.4322 & 0.3080 \\
    $d_\mathrm{merge}=1.5$ & 5.8257 & 1.0760 & 4.1973 & 0.3175 \\
    $d_\mathrm{merge}=2.0$ & 5.8648 & 1.1132 & 4.1006 & 0.3140 \\
    $d_\mathrm{merge}=2.5$ & 6.0023 & 1.0839 & 4.2726 & 0.2961 \\
    \bottomrule
  \end{tabularx}
\end{table}

Table~\ref{tab:sensi} studies the sensitivity of BPDA-GMM to its main hyperparameters: the DP concentration parameter $\alpha_0$, count-decay rate $\lambda$, tempering coefficient $\alpha$, ambiguity threshold $\tau_{\mathrm{amb}}$, and merge distance $d_{\mathrm{merge}}$. Overall, performance is stable around the default configuration. The best trade-off occurs near $\alpha_0=0.5$, $\lambda=0.001$, $\alpha=0.3$, $\tau_{\mathrm{amb}}=0.8$, and $d_{\mathrm{merge}}=1.5$--$2.0$. Too small $\alpha_0$ suppresses new-landmark creation, while too large $\alpha_0$ increases map fragmentation. Similarly, moderate count decay prevents excessive landmark birth as the map grows, and ambiguity-triggered tempering performs best when activated only for sufficiently diffuse posteriors. The merge distance shows a trajectory--map trade-off: larger values improve pose accuracy by merging duplicate landmarks, but overly aggressive merging may reduce semantic map quality.


\section{Conclusion}
\label{sec:conclusion}

We presented \textbf{BPDA-GMM}, an online Bayesian probabilistic data association framework for semantic SLAM with a growing object-level map. By using a Dirichlet-process prior, BPDA-GMM induces a CRP association model that incrementally accumulates association evidence and assigns principled probability mass to new landmarks. CRP-weighted detections update semantic Gaussian landmarks in closed form, forming a Gaussian mixture map whose dominant component is converted into a max-mixture semantic factor. A joint semantic-geometric gate reduces candidate associations, while ambiguity-triggered posterior tempering improves robustness under noisy or inconclusive observations. In the back-end, zero-pose-Jacobian semantic factors decouple landmark refinement from trajectory estimation, preventing noisy semantic detections from directly perturbing the robot pose. Experiments in simulation and on a real indoor dataset show that BPDA-GMM improves trajectory accuracy, semantic mapping quality, and robustness to perceptual aliasing and classifier errors compared with state-of-the-art data association baselines, while maintaining real-time performance on embedded hardware. 

Three limitations point to future work. First, the single-Gaussian landmark model assumes unimodal position uncertainty; elongated objects such as tables or walls would benefit from a mixture of Gaussians per landmark. Second, the semantic indicator gate assumes a fixed discrete class set, and extending to open-vocabulary detectors requires a continuous class embedding. Third, BPDA-GMM does not currently support active data-association planning, in which the robot selects actions to disambiguate hypotheses; incorporating such planning within the CRP framework is a natural open problem. Future work will address these limitations by extending BPDA-GMM to multi-modal object representations, open-vocabulary semantics, active data-association planning, and integrating BPDA-GMM into a multi-robot system.
\bibliographystyle{IEEEtran}
\bibliography{ref}

@article{canh2025semantic,
  title={Semantic Visual Simultaneous Localization and Mapping: A Survey on State of the Art, Challenges, and Future Directions},
  author={Canh, Thanh Nguyen and Zhang, Haolan and HoangVan, Xiem and Chong, Nak Young},
  journal={arXiv preprint arXiv:2510.00783},
  year={2025}
}

@inproceedings{kim2012modeling,
  title={Modeling topic hierarchies with the recursive chinese restaurant process},
  author={Kim, Joon Hee and Kim, Dongwoo and Kim, Suin and Oh, Alice},
  booktitle={Proceedings of the 21st ACM international conference on Information and knowledge management},
  pages={783--792},
  year={2012}
}

@article{bar1975tracking,
  title={Tracking in a cluttered environment with probabilistic data association},
  author={Bar-Shalom, Yaakov and Tse, Edison},
  journal={Automatica},
  volume={11},
  number={5},
  pages={451--460},
  year={1975},
  publisher={Elsevier}
}

@article{reid2003algorithm,
  title={An algorithm for tracking multiple targets},
  author={Reid, Donald},
  journal={IEEE transactions on Automatic Control},
  volume={24},
  number={6},
  pages={843--854},
  year={2003},
  publisher={IEEE}
}

@article{cox1994modeling,
  title={Modeling a dynamic environment using a Bayesian multiple hypothesis approach},
  author={Cox, Ingemar J and Leonard, John J},
  journal={Artificial intelligence},
  volume={66},
  number={2},
  pages={311--344},
  year={1994},
  publisher={Elsevier}
}

@article{murty1968algorithm,
  title={An algorithm for ranking all the assignments in order of increasing cost},
  author={Murty, Katta G},
  journal={Operations research},
  volume={16},
  number={3},
  pages={682--687},
  year={1968},
  publisher={INFORMS}
}

@inproceedings{bowman2017probabilistic,
  title={Probabilistic data association for semantic slam},
  author={Bowman, Sean L and Atanasov, Nikolay and Daniilidis, Kostas and Pappas, George J},
  booktitle={IEEE international conference on robotics and automation (ICRA)},
  pages={1722--1729},
  year={2017},
  organization={IEEE}
}

@inproceedings{doherty2019multimodal,
  author    = {Doherty, Kevin and Fourie, Dehann and
               Leonard, John},
  title     = {Multimodal Semantic {SLAM} with Probabilistic Data
               Association},
  booktitle = {IEEE International Conference on Robotics and
               Automation (ICRA)},
  year      = {2019},
  pages     = {2419--2425},
}

@inproceedings{doherty2020probabilistic,
  author    = {Doherty, Kevin J. and Baxter, David P. and
               Schneeweiss, Edward and Leonard, John J.},
  title     = {Probabilistic Data Association via Mixture Models
               for Robust Semantic {SLAM}},
  booktitle = {IEEE International Conference on Robotics and
               Automation (ICRA)},
  year      = {2020},
  pages     = {1098--1104},
}

@inproceedings{michael2022probabilistic,
  author    = {Michael, Elad and Summers, Tyler and
               Wood, Tony A. and Manzie, Chris and Shames, Iman},
  title     = {Probabilistic Data Association for Semantic {SLAM}
               at Scale},
  booktitle = {IEEE/RSJ International Conference on Intelligent
               Robots and Systems (IROS)},
  year      = {2022},
  pages     = {4359--4364},
}

@inproceedings{wang2018robust,
  author    = {Wang, Jinkun and Englot, Brendan},
  title     = {Robust Exploration with Multiple Hypothesis Data
               Association},
  booktitle = {IEEE/RSJ International Conference on Intelligent
               Robots and Systems (IROS)},
  year      = {2018},
  pages     = {3537--3544},
  address   = {Madrid, Spain},
}

@article{neira2002data,
  title={Data association in stochastic mapping using the joint compatibility test},
  author={Neira, Jos{\'e} and Tard{\'o}s, Juan D},
  journal={IEEE Transactions on robotics and automation},
  volume={17},
  number={6},
  pages={890--897},
  year={2002},
  publisher={IEEE}
}

@inproceedings{hsiao2019mh,
  title={Mh-isam2: Multi-hypothesis isam using bayes tree and hypo-tree},
  author={Hsiao, Ming and Kaess, Michael},
  booktitle={2019 International Conference on Robotics and Automation (ICRA)},
  pages={1274--1280},
  year={2019},
  organization={IEEE}
}

@article{olson2013inference,
  author    = {Olson, Edwin and Agarwal, Pratik},
  title     = {Inference on Networks of Mixtures for Robust Robot
               Mapping},
  journal   = {The International Journal of Robotics Research},
  year      = {2013},
  volume    = {32},
  number    = {7},
  pages     = {826--840},
}

@incollection{brox2021maximum,
  title={Maximum likelihood estimation},
  author={Brox, Thomas},
  booktitle={Computer Vision: A Reference Guide},
  pages={799--801},
  year={2021},
  publisher={Springer}
}

@article{kaess2012isam2,
  title={iSAM2: Incremental smoothing and mapping using the Bayes tree},
  author={Kaess, Michael and Johannsson, Hordur and Roberts, Richard and Ila, Viorela and Leonard, John J and Dellaert, Frank},
  journal={The International Journal of Robotics Research},
  volume={31},
  number={2},
  pages={216--235},
  year={2012},
  publisher={Sage Publications Sage UK: London, England}
}

@techreport{dellaert2012factor,
  author    = {Dellaert, Frank},
  title     = {Factor Graphs and {GTSAM}: A Hands-on Introduction},
  institution={Georgia Institute of Technology},
  number    = {GT-RIM-CP\&R-2012-002},
  year      = {2012},
}

@article{ji2024sgba,
  title={Sgba: Semantic gaussian mixture model-based lidar bundle adjustment},
  author={Ji, Xingyu and Yuan, Shenghai and Li, Jianping and Yin, Pengyu and Cao, Haozhi and Xie, Lihua},
  journal={IEEE Robotics and Automation Letters},
  volume={9},
  number={12},
  pages={10922--10929},
  year={2024},
  publisher={IEEE}
}

@article{ghalamkari2026em,
  author    = {Ghalamkari, Kazu and Hinrich, Jesper L{\o}ve
               and M{\o}rup, Morten},
  title     = {{E$^2$M}: Double Bounded $\alpha$-Divergence
               Optimization for Tensor-based Discrete Density
               Estimation},
  journal   = {Transactions on Machine Learning Research},
  year      = {2026},
}

@article{liu2025slideslam,
  title={Slideslam: Sparse, lightweight, decentralized metric-semantic slam for multirobot navigation},
  author={Liu, Xu and Lei, Jiuzhou and Prabhu, Ankit and Tao, Yuezhan and Spasojevic, Igor and Chaudhari, Pratik and Atanasov, Nikolay and Kumar, Vijay},
  journal={IEEE Transactions on Robotics},
  volume={41},
  pages={6529--6548},
  year={2025},
  publisher={IEEE}
}

@article{yang2019cubeslam,
  author  = {Yang, Shichao and Scherer, Sebastian},
  title   = {{CubeSLAM}: Monocular {3D} Object {SLAM}},
  journal = {IEEE Transactions on Robotics},
  volume  = {35},
  number  = {4},
  pages   = {925--938},
  year    = {2019}
}

@article{nicholson2019quadricslam,
  author  = {Nicholson, Lachlan and Milford, Michael and S{\"u}nderhauf, Niko},
  title   = {{QuadricSLAM}: Dual Quadrics from Object Detections as Landmarks in Object-Oriented {SLAM}},
  journal = {IEEE Robotics and Automation Letters},
  volume  = {4},
  number  = {1},
  pages   = {1--8},
  year    = {2019}
}

@inproceedings{qian2021semantic,
  author    = {Qian, Zhentian and Patath, Kartik and Fu, Jie and Xiao, Jing},
  title     = {Semantic {SLAM} with Autonomous Object-Level Data Association},
  booktitle = {IEEE International Conference on Robotics and Automation (ICRA)},
  pages     = {11203--11209},
  year      = {2021}
}

@inproceedings{wang2021dspslam,
  author    = {Wang, Jingwen and R{\"u}nz, Martin and Agapito, Lourdes},
  title     = {{DSP-SLAM}: Object Oriented {SLAM} with Deep Shape Priors},
  booktitle = {International Conference on 3D Vision (3DV)},
  pages     = {1362--1371},
  year      = {2021}
}

@inproceedings{wu2020eao,
  author    = {Wu, Yanmin and Zhang, Yunzhou and Zhu, Delong and Feng, Yonghui and Coleman, Sonya and Kerr, Dermot},
  title     = {{EAO-SLAM}: Monocular Semi-Dense Object {SLAM} Based on Ensemble Data Association},
  booktitle = {IEEE/RSJ International Conference on Intelligent Robots and Systems (IROS)},
  pages     = {4966--4973},
  year      = {2020}
}

@article{wu2023object,
  title={An object slam framework for association, mapping, and high-level tasks},
  author={Wu, Yanmin and Zhang, Yunzhou and Zhu, Delong and Deng, Zhiqiang and Sun, Wenkai and Chen, Xin and Zhang, Jian},
  journal={IEEE Transactions on Robotics},
  volume={39},
  number={4},
  pages={2912--2932},
  year={2023},
  publisher={IEEE}
}

\end{document}